\documentclass{article}
\usepackage{ijcai21}

\usepackage{times}
\usepackage{soul}
\usepackage{url}
\usepackage[hidelinks]{hyperref}
\usepackage[utf8]{inputenc}
\usepackage[small]{caption}
\usepackage{graphicx}
\usepackage{amsmath}
\usepackage{amsthm}
\usepackage{booktabs}
\usepackage{algorithm}
\usepackage{algorithmic}
\usepackage{subfigure}
\usepackage{multirow}
\usepackage[usenames,dvipsnames,svgnames,table]{xcolor}

\title{MathBERT: A Pre-Trained Model for Mathematical Formula Understanding}

\author{
    Shuai Peng, Ke Yuan, Liangcai Gao, Zhi Tang
    \affiliations
    Peking University
    \emails
    $\left\{pengshuaipku,yuanke,gaoliangcai,tangzhi\right\}@pku.edu.cn$
}

\begin{document}

\maketitle

\begin{abstract}
Large-scale pre-trained models like BERT, have obtained a great success in various Natural Language Processing (NLP) tasks, while it is still a challenge to adapt them to the math-related tasks. Current pre-trained models neglect the structural features and the semantic correspondence between formula and its context. To address these issues, we propose a novel pre-trained model, namely \textbf{MathBERT}, which is jointly trained with mathematical formulas and their corresponding contexts. In addition, in order to further capture the semantic-level structural features of formulas, a new pre-training task is designed to predict the masked formula substructures extracted from the Operator Tree (OPT), which is the semantic structural representation of formulas. We conduct various experiments on three downstream tasks to evaluate the performance of MathBERT, including mathematical information retrieval, formula topic classification and formula headline generation. Experimental results demonstrate that MathBERT significantly outperforms existing methods on all those three tasks. Moreover, we qualitatively show that this pre-trained model effectively captures the semantic-level structural information of formulas. To the best of our knowledge, MathBERT is the first pre-trained model for mathematical formula understanding.
  
\end{abstract}

\section{Introduction} \label{section:introduction}

Mathematical formulas are widely used in the fields of science, technology and engineering. Several research tasks on mathematical formula, including Mathematical Information Retrieval(MIR)~\cite{yuan2016mathematical,TangentS,TangentCFT}, Mathematical Formula Understanding (MFU)~\cite{jiang2018mathematics,yuanke} and so forth, have continuously attracted researchers’ attention. Processing mathematical information is still a challenging task due to the diversity of mathematical formula representations, the complexity of formula structure and the ambiguity of implicit semantics. Researchers utilize non-pretrained customized models to solve specific math-related tasks. They are built upon either the structural features of formula~\cite{TangentCFT} or topical correspondence between formula and context~\cite{Topiceq}, but do not consider a joint training of structural and semantic information. In the past decades, large-scale pre-trained models such as ELMo~\cite{ELMo}, GPT~\cite{GPT}, BERT~\cite{BERT} and XLNet~\cite{XLNet} have achieved great advancement on various Natural Language Processing (NLP) tasks. The success in NLP also drives the development of pre-trained model in other specific fields such as VideoBERT~\cite{VideoBERT} for video, CodeBERT~\cite{codebert} for code, LayoutLM~\cite{LayoutLM} for document. Inspired by the success of these pre-trained models, we assume the pre-trained model will also benefit the math-related research.

\begin{figure}[t]
\vskip -0.2in
\begin{center}
\centerline{\includegraphics[width=\columnwidth]{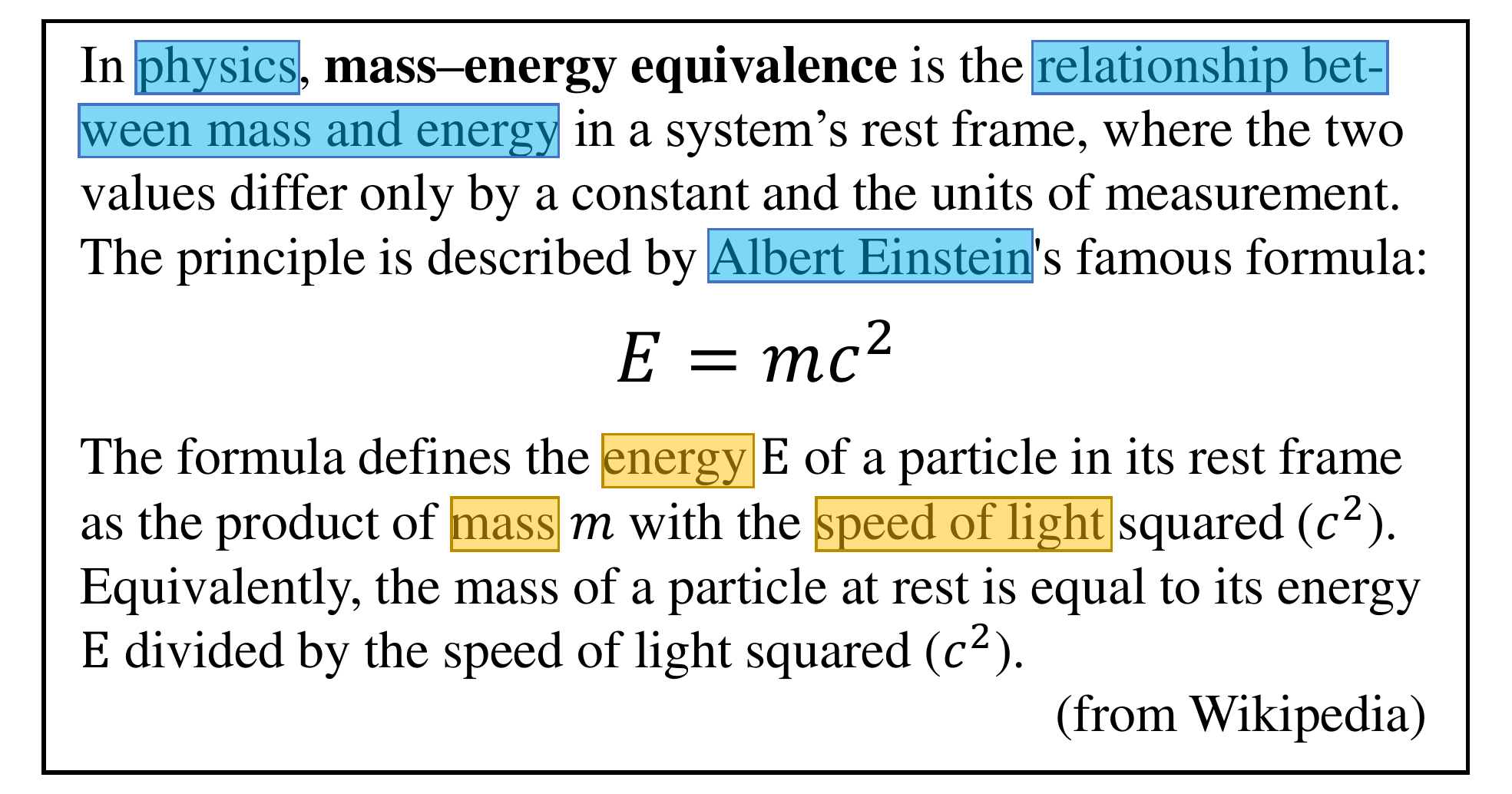}}
\vskip -0.1in
\caption{An example of mathematical formula ``$E=mc^2$" with its context, where the text contains rich semantic information of the brief formula.}
\label{figure-wikipedia_demo}
\end{center}
\vskip -0.35in
\end{figure}

Intuitively, formula is not only a simple sequence of mathematical symbols but also has a strong semantic relation with its context, as is illustrated in Figure \ref{figure-wikipedia_demo}. The available information from the single formula is limited. For instance, we merely acquire an equation that $E$ is equal to $m$ times $c$ squared. Much more semantic information that is vital for formula understanding is often included in its context, such as the meaning of each symbol ($E$ for ‘energy’, $m$ for ‘mass’, $c$ for ‘light speed’), as well as some significant associated information of the formula, including its domain (physics), its name (mass-energy equivalence), its inherent meaning (the relationship between mass and energy) and even its proposer (Albert Einstein). Therefore, to fully exploit the complementary relationship between formula and context, MathBERT is jointly trained with formula and its context. Two pre-training tasks are employed to learn representations of formula which are \textit{Masked Language Modeling} (MLM) and \textit{Context Correspondence Prediction} (CCP). Furthermore, mathematical formula contains rich structural information, which is important to semantic understanding and formula retrieval tasks. Thus, we take the Operator Trees (OPTs) as the input and design a novel pre-training task named \textit{Masked Substructure Prediction} (MSP) to capture semantic-level structural information of formula. 

Furthermore, We build a large dataset containing more than 8.7 million formula-context pairs which are extracted from scientific articles published on arXiv.org\footnote{\url{https://arxiv.org}} and train MathBERT on it. The model is evaluated on three downstream tasks, including mathematical information retrieval, formula topic classification and formula headline generation. Experimental results demonstrate that MathBERT significantly outperforms existing methods on all three tasks. Moreover, we qualitatively show that the proposed model could effectively capture the semantic-level structural information of formulas. 

The main contributions of this work are summarized as follows:
\begin{itemize}
    \item The first pre-trained model for mathematical formula understanding is proposed, which is jointly trained with formulas, contexts and OPTs.
    \item A novel pre-training task is designed to capture the semantic-level structural information of formulas.
    \item The proposed MathBERT model achieves a significant improvement compared with the strong baselines on all three downstream tasks.
    \item A new dataset for formula topic classification is constructed, which contains mathematical formulas and their corresponding contexts, and would be open soon.
\end{itemize}

\section{Related Work}
In this section, we describe the related works from the Pre-trained Models to the Mathematical Formula Representation.


\subsection{Pre-Trained Models}

Pre-tained model obtained an increasing attention since the great successes were achieved in a variety of NLP tasks, such as ELMo~\cite{ELMo}, GPT~\cite{GPT}, BERT~\cite{BERT}, XLNet~\cite{XLNet}. These pre-trained models performed well in general NLP tasks like text classification~\cite{BERT}, machine translation~\cite{BERT-fused,sundararaman2019syntax} and machine summarization~\cite{miller2019leveraging,xenouleas2019sumqe}. However, these models were not good at dealing with the specific objects. Thus some specific pre-trained models were proposed. For instance, CodeBERT~\cite{codebert} is a pre-trained model for the code synthesis which was jointly trained on programming and natural languages. LayoutLM~\cite{LayoutLM} was proposed for document understanding, which was jointly trained on multi-modal information including text, image and layout. 

\subsection{Mathematical Formula Representation}

\begin{figure}[h]
    \vskip -0.1in
    \centering
    \subfigure[]{
        \includegraphics[width=0.7in]{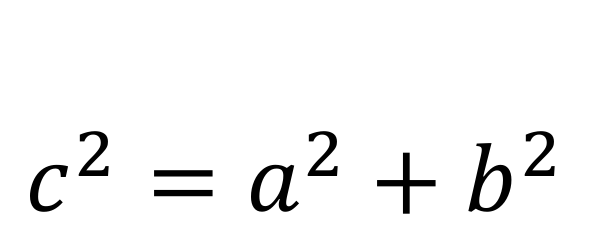}
    }
    \subfigure[]{
        \includegraphics[width=1.2in]{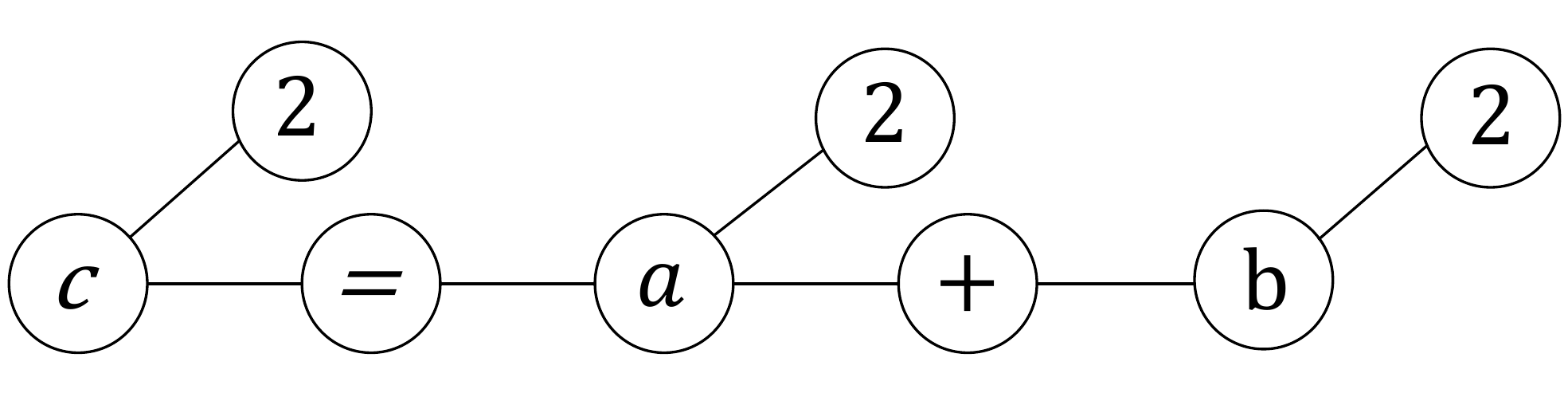}
    }
    \subfigure[]{
        \includegraphics[width=1.2in]{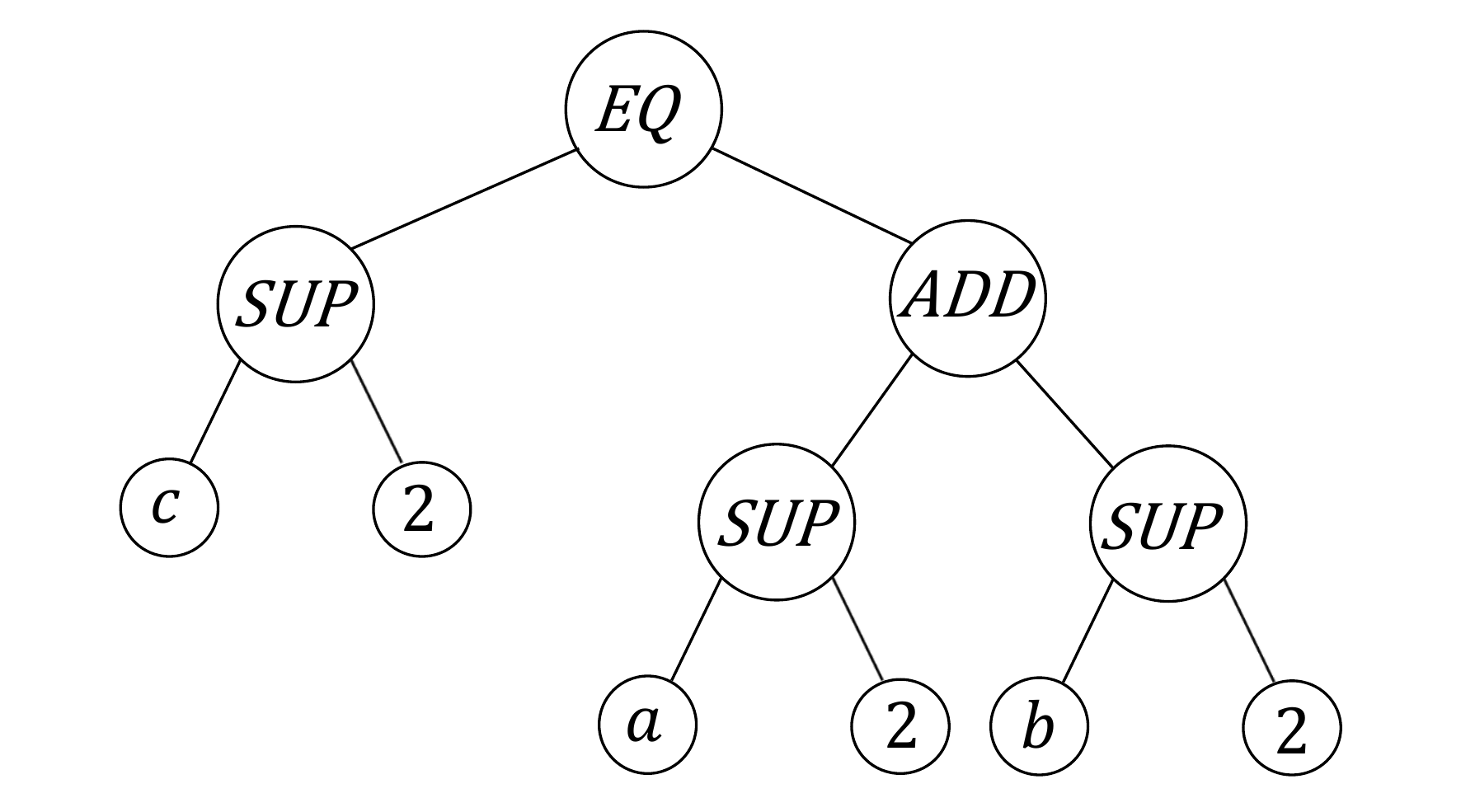}
    }
    \vskip -0.1in
    \caption{Formula (a) $c^2=a^2+b^2$ with its Symbol Layout
Tree (SLT) (b), and Operator Tree (OPT) (c). SLTs represent formula appearance by the spatial arrangements of math symbols, while OPTs
define the mathematical operations represented in expressions.}
    \label{fig:slt_opt}
    \vskip -0.1in
\end{figure}

The representation of mathematical formulas is important to the math-related tasks, such as mathematical information retrieval~\cite{wang2015wikimirs,yuan2016mathematical,TangentS,jiang2018mathematics,TangentCFT} and math expression generation~\cite{yuanke,zhang2020tree}. Some works treat mathematical formulas as a sequence of symbols and use the one-hot representations~\cite{Topiceq,yuan2016mathematical}. However, distinct from plain text, mathematical formulas contain strong structural features~\cite{TangentCFT,yuanke}. Thus some works~\cite{wang2015wikimirs,yuan2016mathematical,jiang2018mathematics,TangentS,TangentCFT} utilized the tree structure to represent mathematical formulas, including the Symbol Layout Tree (SLT) and Operator Tree (OPT). For instance, two different tree representations of the formula ``$c^2=a^2+b^2$" are shown in the Figure~\ref{fig:slt_opt}. In this work, OPT is selected as the input of MathBERT rather than SLT based on the following two considerations. First, layout information of formula in SLT has been included in \LaTeX{} codes to some extent. Second, and most important, OPT plays a crucial role in incorporating semantic-level structural information for the reason that it contains mathematical syntax and semantics which guides the recovery of mathematical operations~\cite{zanibbi2012recognition}.

\begin{figure*}[t]
    \centering
    \resizebox{0.94\linewidth}{!}{
    \includegraphics{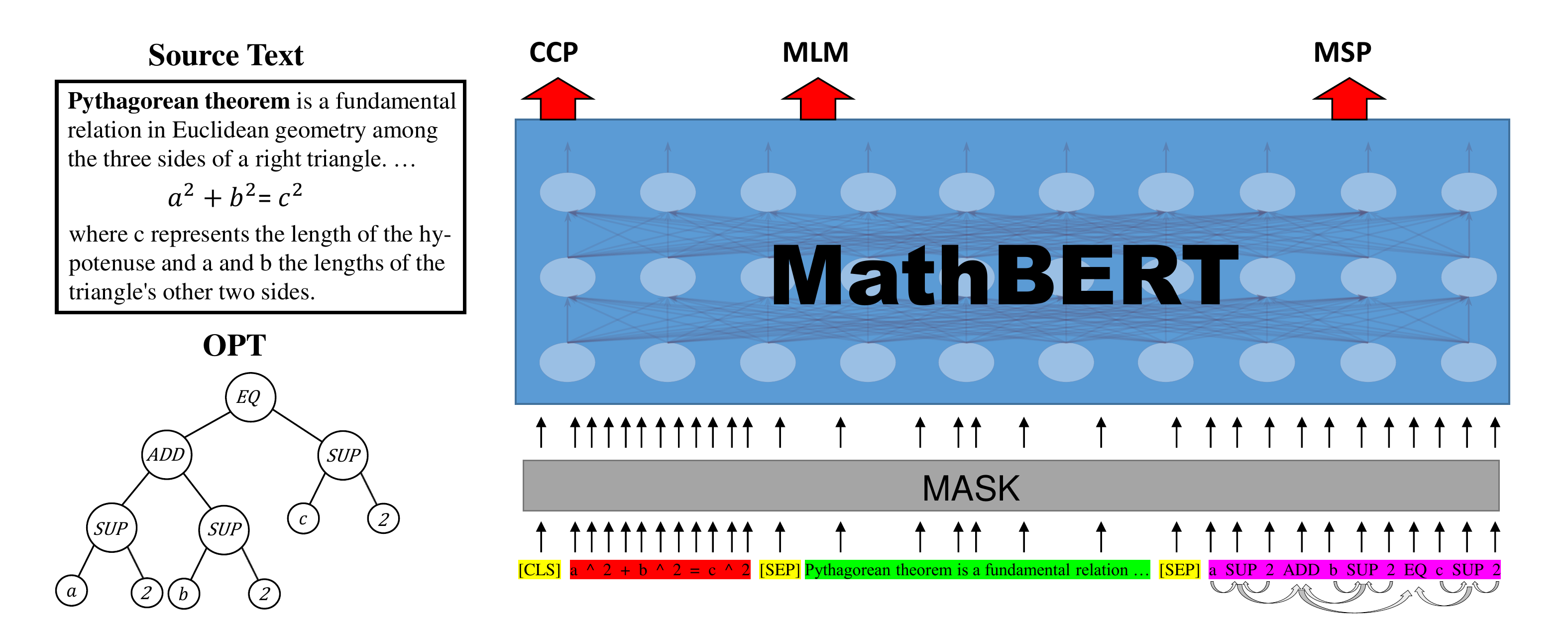}
    }
    \vskip -0.1in
    \caption{An illustration of the architecture of MathBERT.
    The two figures on the left indicate the source text extracted from scientific articles which consists of mathematical formula and its context, and the associated OPT translated from \LaTeX{} code of the formula. Raw text is tokenized and concatenated with \LaTeX{} tokens and operators as the input. In the pre-training stage, we randomly mask the input and employ three pre-training tasks (MLM,CCP,MSP) to train MathBERT. To learn structure-aware information of formula, we utilize the structure of OPT to modify attention mask matrix in Transformers and train MathBERT with MSP pre-training task.}
    \label{fig:architecture}
\end{figure*}

\section{MathBERT}

In this section, we introduce our proposed MathBERT, including the model architecture, pre-training tasks, pre-training data and pre-training details.

\subsection{Model Architecture}

An enhanced multi-layer bidirectional Transformer~\cite{Transformer} is built as the backbone of MathBERT, which is modified from vanilla BERT. Considering that there is much implicit semantic information hidden in the context and structural information implied by formula, we concatenate the formula \LaTeX{} tokens, context and operators together as the input of MathBERT. Moreover, the attention mechanism in Transformer is modified based on the structure of OPT to enhance its ability of capturing structural information. The overall architecture of MathBERT is shown in Figure \ref{fig:architecture}.

Given a sequence of LaTeX tokens $ T $ = \{$ t_1$, $t_2$, …,$t_{L_T}$\}, its context  $ C $ = \{$c_1$, $c_2$,…,$c_{L_C}$\} and its operator tree $ OPT=(N, E) $ where $ N$=\{$n_1$, $n_2$, …, $n_{L_N}$\} is the set of operators, $ E $= \{$e_1$, $e_2$, … ,$e_{L_E}$\} is the set of edges, we set the input as the concatenation of the above three, that is $ [CLS]$, $t_1$, $t_2$, …, $t_{L_T}$, $[SEP]$, $c_1$, $c_2$, …, $c_{L_C}$, $[SEP]$, $n_1$, $n_2$, …, $n_{L_N}$. Here $ [CLS] $ is a special classification token whose final hidden vector is often considered as the aggregate sequence representation for classification tasks, and $ [SEP] $ is a special token used to separate the three segments.

\begin{figure}[!t]
\vskip -0.25in
\begin{center}
\centerline{\includegraphics[scale=0.6]{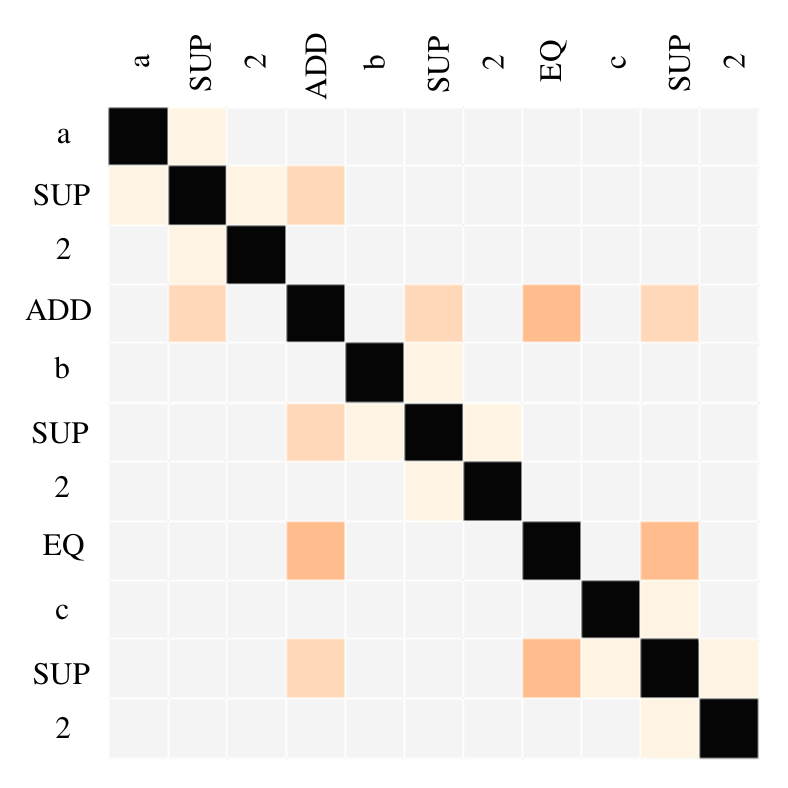}}
\vskip -0.1in
\caption{An illustration of modified attention mask map with the input of $a^2+b^2=c^2$. Gray squares denote that there is no edge between these two operators so we mask them by 0, which results in a consequence that their attention weights go to -$\infty$. Orange squares denote there exists an edge between them and black squares mean they are the same node. Attention is applied as normal in these two cases.}
\label{figure-attenion_mask}
\end{center}
\vskip -0.4in
\end{figure}

In order to explicitly incorporate semantic-level structural information from OPT, we do not simply follow BERT which treats the operators as other normal tokens to attend them together densely in attention mechanism. Instead, the edges between operators are leveraged to modify the attention mask matrix, as is illustrated in Figure \ref{figure-attenion_mask}. For any two different nodes $n_i$ and $n_j$, if there does not exist an edge $ e_k \in E $ between them, the corresponding values $M_{(i,j)}$ and $M_{(j,i)}$ in attention mask matrix $M$ are masked by $0$ to avoid the two nodes to attend each other directly while the other values in $M$ remain $1$. Formally, it can be represented as follows: 
\begin{equation}
    M_{(i,j)} = 
    \begin{cases}
        0 & \text{if $\langle n_i,n_j \rangle \notin E$ and $\langle n_j,n_i \rangle \notin E$ and $i \neq j$}. \\
        1 & \text{otherwise}.
    \end{cases}
\end{equation}

\subsection{Pre-Training Tasks}

We expect MathBERT to obtain three aspects of information: text representations, latent relationship between formula and context, and semantic-level structure of formula, which correspond to the following three pre-training tasks respectively.

\subsubsection{Masked Language Modeling}

\textit{Masked Language Modeling} is presented in BERT~\cite{BERT} to address the problem of ‘see itself’ in traditional bidirectional language modeling, which has been proved effective to learn text representations. Concretely, given the input $[CLS]$ $T$ $[SEP]$ $C$ $[SEP]$ $N$, 15 $\%$ of tokens $T_{mask}$ and $C_{mask}$ are randomly sampled from $T$ and $C$ for masking operation, in which 80 $\%$ of them are replaced with $[MASK]$, 10 $\%$ of them are randomly replaced by other arbitrary tokens, and 10 $\%$ of them remain unchanged. The objective is to predict the original tokens which are masked out, formulated as follows:
\begin{equation}
    Loss_{MLM} = \sum_{x_i \in T_{mask} \cup C_{mask}} -\log p(x_i)
\end{equation}
where $p(x_i)$ denotes the probability of predicting the original token correctly in the position of $x_i$. Particularly, owing to the complementary relationship among formula, context and operators, it is encouraged to utilize the information from other segments to predict the masked tokens, which contributes to establishing connections among the three segments.

\subsubsection{Context Correspondence Prediction}

As mentioned in Section~\ref{section:introduction}, there is a latent semantic relation between mathematical formula and its context, which is not directly captured by language modeling. Therefore, similar to the \textit{Next Sentence Prediction} task in BERT, we pre-train for a binarized \textit{Context Correspondence Prediction} task. Specifically, 50 $\%$ of context $C$ in pre-training examples are randomly replaced with another context in the dataset. The objective is to predict whether the current input of context $C^{\prime}$ is the corresponding context of $T$ or not, which can be formulated as follows, where $p$ denotes the probability of $C=C^{\prime}$:
\begin{align}
    Loss_{CCP} = -\delta \log p - (1-\delta) \log (1-p)
\end{align}
\vskip -0.15in
\begin{equation}
    \delta=\begin{cases}
        1 & \text{if $C = C^{\prime}$}. \\
        0 & \text{otherwise}.
    \end{cases}
\end{equation}

\subsubsection{Masked Substructure Prediction}

In order to incorporate structural information from OPTs, we present a pre-training task named \textit{Masked Substructure Prediction}. Substructure here means the structure composed of an operator, its parent node and child nodes as a part of the OPT. In practice, 15 $\%$ of nodes $N_{mask}$ are randomly sampled from the input $N$. For every node $n_i$ in $N_{mask}$, we cut off all the connections with its parent node and child nodes to mask the substructure which $n_i$ belongs to. The objective is to predict the parent node and child nodes of the masked $n_i$, formulated as follows, where $p(n_i,n_j)$ denotes the probability that $n_j$ is the parent or child node of $n_i$.
\begin{align}
   Loss_{MSP} = & \sum_{n_i \in N_{mask}} \sum_{n_j \in N} \bigg( -\delta \log p(n_i,n_j) \nonumber\\ & - (1-\delta) \log \big(1-p(n_i,n_j)\big) \bigg)
\end{align}
\vskip -0.1in
\begin{equation}
    \delta=\begin{cases}
        1 & \text{if $e_{i,j} \in E$ or $e_{j,i} \in E$} \\
        0 & \text{otherwise}
    \end{cases}
\end{equation}

The total loss is calculated by simply adding the above three together:
\begin{equation}
    Loss_{total} = Loss_{MLM} + Loss_{CCP} + Loss_{MSP}
\end{equation}

\subsection{Pre-Training Data}\label{subsection:pre-training_data}

Since it is the first pre-trained model for mathematical formulas, there is scarcely a large public dataset that consists of formula-context pairs. As such, we build the pre-training dataset with the public scientific articles from arXiv.org. Arxiv bulk data available from Amazon S3\footnote{\url{https://arxiv.org/help/bulk_data_s3}} is the complete set of arxiv documents which contains source TEX files and processed PDF files. “$\backslash$begin\{equation\} … $\backslash$end\{equation\}” is used as the matching pattern to extract single-line display formulas from \LaTeX{} source in these TEX files. We collect the surrounding text with at least 400 characters as the context of formula and replace the formula with a special token $[MATH]$ to indicate the position. As for data-preprocessing, we utilize the toolkit \LaTeX{} tokenizer in im2markup\footnote{\url{https://github.com/harvardnlp/im2markup}} to tokenize separately formulas and OPT translator in TangentS\footnote{\url{https://github.com/BehroozMansouri/TangentCFT/tree/master/TangentS}} to convert \LaTeX{} codes into OPTs. Finally, we obtain a large dataset that consists of 8.7 million formulas with contexts and corresponding OPTs.

\subsection{Pre-Training Details}

We train MathBERT on 4 NVIDIA TITAN X 12GB GPUs with total batch size of 48. To well utilize the existing pre-trained model in NLP and accelerate the training process, we initialize the weights of MathBERT with the pre-trained BERT base model released by Google\footnote{\url{https://github.com/google-research/bert}} which has a 12-layer Transformer with 768 hidden sizes. Due to the limitation of GPU memory, the max length of input sequences is set as 256. The Adam optimizer is used with the learning rate of 2e-5. It costs two weeks to train MathBERT on 8.7M data with around 10,000,000 iterations.

\section{Experiment}

To verify the effectiveness of MathBERT, we conduct experiments and evaluate it on three downstream tasks: mathematical information retrieval, formula topic classification and formula headline generation. Additionally, ablation study is done followed by qualitative analysis, indicating that MathBERT well captures the semantic-level structural information of formulas.

\subsection{Mathematical Information Retrieval}\label{subsection:MIR}

\begin{table}[!h]
\vskip -0.05in
\centering
\begin{tabular}{lccc}
\toprule
Approaches         & Partial & Full & H-Mean \\
\hline
MCAT         & 56.98             & 56.78         & 56.88        \\
TangentS     & 58.72             & 63.61         & 61.07        \\
Approach0    & 59.50             & 67.26         & 63.14        \\
TangentCFT   & 71.34             & 59.63         & 64.96        \\
\hline
BERT         & 70.53             &	58.33         & 63.85        \\
\textbf{MathBERT}           & 73.61             & 61.35         & 66.92   
\\
\textbf{MathAPP} & \textbf{76.07}             & \textbf{71.61}         & \textbf{73.77}  
  \\
\bottomrule
\end{tabular}
\vskip -0.05in
\caption{NTCIR-12 Results (Avg. \textit{bpref}@1000). H-Mean denotes the harmonic mean of partial relevance and full relevance score.}
\label{tab:MathIR}
\vskip -0.1in
\end{table}

\begin{table*}[!tbp]
\begin{center}
\begin{tabular}{lcccccc}
\toprule
\multirow{2}{*}{Models}              & \multicolumn{3}{c}{Formula Only} & \multicolumn{3}{c}{Formula   with Context} \\
        & Precision    & Recall   & F1       & Precision      & Recall      & F1          \\
\hline            
TextRNN    & 56.86       & 56.87   & 56.63   & 64.75         & 64.22      & 64.33      \\
TextRNN\_Att      & 57.72       & 57.30   & 57.30   & 65.38         & 65.21      & 65.15      \\
TextRCNN          & 58.92       & 58.78   & 58.18   & 65.04         & 65.11      & 64.82      \\
FastText        & 60.04       & 59.84   & 59.82   & 68.21         & 68.08      & 68.03      \\
\hline
BERT            & 60.82       & 60.02   & 60.34   & 71.55         & 70.29      & 70.84      \\
\textbf{MathBERT}          & \textbf{65.18}       & \textbf{64.05}   & \textbf{64.52}   & \textbf{75.68}         & \textbf{74.46}      & \textbf{75.03}     \\
\bottomrule
\end{tabular}
\end{center}
\vskip -0.1in
\caption{TopicMath-100K Results, evaluated with macro-average precision, recall and F1 score on 10 classes. Formula only and formula with context are respectively used as the input.}
\label{tab:classification}
\vskip -0.1in
\end{table*}

\begin{table}[!t]
\begin{center}
\begin{tabular}{lc}
\toprule
Class & Data   \\
\midrule
Astrophysics & 6,426 \\
Machine Learning & 7,597 \\
Theoretical Economics & 11,442 \\
Relativity & 19,386 \\
High Energy Physics Theory & 20,856 \\
Number Theory & 18,954 \\
Nuclear Theory & 11,262 \\
Atomic Physics & 7,279 \\
Computational Finance & 13,035 \\
Quantum Physics & 16,065 \\
All      & 132,302 \\
\bottomrule
\end{tabular}
\end{center}
\vskip -0.1in
\caption{Statistics of TopicMath-100K.}
\label{tab:TopicMathData}
\vskip -0.15 in
\end{table}

Similar to other information retrieval (IR) tasks, given a formula as the query, mathematical information retrieval aims to return the relevance of formulas in a large set of documents. Formulas can be indexed using vector similarity measures for retrieval. Hence, it is a suitable downstream task to evaluate the output embeddings of MathBERT. 
Here MathBERT is evaluated on the NTCIR-12 MathIR Wikipedia Formula Browsing Task~\cite{NTCIR-12}, which is the most current benchmark for formula retrieval. The dataset contains over 590,000 mathematical formulas from English Wikipedia and 20 non-wildcards queries. There are two human assessors evaluating the pooled hits from participating system by scoring the hit with the score 2, 1 or 0 from highly relevant to irrelevant. The final hit relevance rating is the sum of the two assessor scores (from 0 to 4), with scores of 3 or higher considered fully relevant and other scores of 1 or higher considered partially relevant. We regard the mean of the last two layers’ feature vectors in MathBERT as formula embeddings and reorder the top-1000 results of TangentCFT~\cite{TangentCFT} according to cosine similarity over formula vectors. Then we use \textit{bpref} as the metric to compare our results with previous approaches, including MCAT~\cite{MCAT}, TangentS~\cite{TangentS}, Approach0~\cite{Approach0} and TangentCFT. The results are shown in Table \ref{tab:MathIR}.

MathBERT achieves the highest partial and harmonic mean \textit{bpref} score. Due to the lack of mathematical and structure-aware information, BERT pre-trained on NLP data obtains a poor result on this task. Compared with another embedding model TangentCFT, our reordered results outperform its original top-1000 results on all the metrics. However, the results on full \textit{bpref} score are still lower than TangentS and Approach0, which may be explained by the shortage of using cosine similarity over formula vectors rather than using direct comparison of formula trees. Consequently, we follow the approach in TangentCFT and create another model (\textit{MathAPP}) by combining retrieval scores from MathBERT and Approach0, achieving state-of-the-art performance.

\begin{table}[!t]
\begin{center}
\setlength{\tabcolsep}{1.4mm}
\begin{tabular}{lccccc}
\toprule
            & R1    & R2    & RL    & BLEU-4 & METEOR \\
\hline        
Random      & 31.56 & 21.35 & 28.99 & 24.32  & 23.40 \\
Tail        & 22.55 & 14.69 & 20.76 & 22.23  & 23.78 \\
Lead        & 42.23 & 31.30 & 39.29 & 29.89  & 31.61 \\
TextRank    & 42.19 & 30.85 & 38.99 & 28.29  & 31.78 \\
\hline
Seq2Seq     & 52.14 & 38.33 & 49.00 & 42.20  & 30.65 \\
PtGen       & 53.26 & 39.92 & 50.09 & 44.10  & 31.76 \\
Transformer & 54.49 & 40.57 & 50.90 & 45.79  & 32.92 \\
\hline
BERT-fused        & 60.76 & 46.98 & 51.74 & 47.08  & 33.46 \\
\textbf{MathBERT}    & \textbf{61.25}  & \textbf{48.06}  & \textbf{57.72} & \textbf{49.40}   & \textbf{34.67}   \\
\bottomrule
\end{tabular}
\end{center}
\vskip -0.1in
\caption{EXEQ-300K Results, evaluated with F1 scores of R1 (ROUGE-1), R2 (ROUGE-2), RL (ROUGE-L), BLEU-4 and METEOR.}
\label{tab:generation}
\vskip -0.2 in
\end{table}

\subsection{Formula Topic Classification}

Formula topic classification is a typical multi-class classification task like text classification in NLP, where the goal is to predict which topic a mathematical formula belongs to. Following the approach described in Section~\ref{subsection:pre-training_data}, we collect 132,302 formula-context pairs from scientific articles published on arXiv.org within a year in 10 selected topics as our dataset named TopicMath-100K. Data statistics is shown in Table \ref{tab:TopicMathData}. TopicMath-100K is randomly split into train (80 $\%$ , 105,841), validation (10 $\%$ , 13,230) and test (10 $\%$ , 13,231) sets. We conduct experiments on this dataset and compare our results with several non-pretrained models and BERT. The results are shown in Table \ref{tab:classification}.

\begin{table*}[!t]
\begin{center}
\begin{tabular}{lccccccccc}
\toprule
\multirow{3}{*}{Settings} & \multicolumn{3}{c}{NTCIR-12} & \multicolumn{6}{c}{TopicMath-100K}  \\
& \multirow{2}{*}{Partial} & \multirow{2}{*}{Full} & \multirow{2}{*}{H-Mean} & \multicolumn{3}{c}{Only   Formula} & \multicolumn{3}{c}{Formula   with Context} \\
&&&& Precision    & Recall   & F1       & Precision      & Recall      & F1    \\
\hline
\textbf{MathBERT}          & \textbf{73.61}                              & \textbf{61.35}                          & \textbf{66.92}                         & \textbf{65.18}       & \textbf{64.05}   & \textbf{64.52}   & \textbf{75.68}         & \textbf{74.46}      & \textbf{75.03}      \\
-w/o OPT          & 72.84                              & 61.05                          & 66.43                         & 64.80       & 63.57   & 64.10   & 75.24         & 73.72      & 74.38      \\
-w/o context   & 73.24 & 60.92 & 66.51 &	64.65 & 63.51 & 64.01 &	73.42 & 73.01 & 73.17 \\
-w/ formula only  & 72.36                              & 60.35                          & 65.81                         & 64.67       & 63.44   & 63.97   & 73.36         & 72.91      & 73.11  \\   
\bottomrule
\end{tabular}
\end{center}
\vskip -0.15in
\caption{Results on NTCIR-12 and TopicMath-100K with different pre-training settings.}
\label{tab:ablation}
\vskip -0.1 in
\end{table*}

\begin{table}[!t]
\centering
\begin{tabular}{rlc}
\toprule
\multicolumn{3}{c}{MathBERT}\\
Rank & Formula  &	Similarity \\
\hline
1 & $\frac{a+b}{c+d}$ &	1.0 \\
2 & $(a+b)/(c+d)$ &	0.9636 \\
3 & $(a+b)\div(c+d)$ & 0.9447 \\
4 & $(a+b)\times(c+d)$ & 0.9251 \\
5 & $\frac{1+2}{3+4}$ & 0.9248 \\
6 & $\frac{5+6}{7+8}$ & 0.9005 \\
… & … & … \\
\hline\hline
\multicolumn{3}{c}{MathBERT -w/o OPT}\\
Rank & Formula  &	Similarity \\
\hline
1 & $\frac{a+b}{c+d}$ &	1.0 \\
2 & $\frac{1+2}{3+4}$ & 0.9143 \\
3 & $(a+b)/(c+d)$ &	0.9130 \\
4 & $\frac{5+6}{7+8}$ & 0.8923 \\
5 & $(a+b)\div(c+d)$ & 0.8680 \\
6 & $(a+b)\times(c+d)$	& 0.8594 \\
… & … & … \\
\hline\hline
\multicolumn{3}{c}{BERT}\\
Rank & Formula  &	Similarity \\
\hline
1 & $\frac{a+b}{c+d}$ &	1.0 \\
2 & $\frac{1+2}{3+4}$ & 0.9036 \\
3 & $\frac{5+6}{7+8}$ & 0.8770 \\
4 & $(a+b)\times(c+d)$ & 0.8526 \\
5 & $(a+b)\div(c+d)$ & 0.8165 \\
6 & $(1+2)\times(3+4)$ & 0.7529 \\
… & … & … \\
\bottomrule
\end{tabular}
\vskip -0.1in
\caption{The ranking results according to the cosine similarity with $\frac{a+b}{c+d}$.}
\label{tab:qualitative}
\vskip -0.15in
\end{table}

MathBERT achieves state-of-the-art performance on all metrics, especially outperforms vanilla BERT significantly. Taking only formula as input, BERT pre-trained on natural language data does not obtain a much better result than those non-pretrained models, which implies that pre-training model on mathematical formulas can improve formula topic classification indeed.

\subsection{Formula Headline Generation}

Formula headline generation is a summarization task aiming to generate a concise math headline from a detailed math question which contains math formulas and descriptions. Here we use EXEQ-300K proposed in \cite{yuanke} as the dataset and conduct experiments to investigate the performance of MathBERT on generation tasks. Specifically, following BERT-fused \cite{BERT-fused}, we utilize MathBERT to extract representations for an input sequence, and fuse them with each layer of the encoder and decoder of Transformer through attention mechanism to generate the headline. The obtained results are compared with four extractive methods (Random, Tail, Lead and TextRank) and four abstractive methods (Seq2Seq~\cite{LSTM}, PtGen~\cite{PtGen}, Transformer~\cite{Transformer} and BERT-fused~\cite{BERT-fused}). The results are shown in Table \ref{tab:generation}. MathBERT outperforms other models on all evaluation metrics, especially Transformer and BERT-fused, which implies that the formula and context representations from MathBERT contribute to downstream generation model.

\subsection{Ablation Study}

To explore the impact of different modalities and pre-training tasks, an ablation study is conducted on mathematical information retrieval task and formula topic classification task, respectively. Four different pre-training settings are applied in experiments: 1) using formula, context and OPT as inputs and all three pre-training tasks, 2) without OPT and MSP pre-training task, 3) without context and CCP pre-training task, 4) with only formula and MLM pre-training task. The results of different settings are shown in Table \ref{tab:ablation}.

Pre-training model using only formula as pre-training input always leads to the lowest results. The effects of pre-training with context or OPT vary with the different downstream tasks. Specifically, OPT contributes more to IR task that is sensitive to formula structure, while context is more important in topic classification which concerns inherent meaning of formula. 

\subsection{Qualitative Analysis}

To demonstrate the effectiveness of MathBERT in learning semantic-level structural information of mathematical formulas, we further conduct qualitative analysis. Concretely, 15 formulas containing similar symbols are selected, some of which have equative meanings in mathematics. Following the approach in Section~\ref{subsection:MIR}, we employ three embedding models to extract feature vectors from these formulas and rank them by cosine similarities. The results are shown in Table \ref{tab:qualitative}. 

As the results indicate, BERT only considers the similarity of appearance, resulting in the poor ranking between $ (a+b)\times(c+d) $ and $ (a+b)\div(c+d) $. Without OPT input, the embeddings of MathBERT still retain some semantic information, which can be proved by the increased ranking of $ (a+b)/(c+d) $ and $ (a+b)\div(c+d) $ . As observed from the result of MathBERT, $ (a+b)/(c+d) $ and $ (a+b)\div(c+d) $ are two of the most similar formulas to $ \frac{a+b}{c+d} $, which demonstrates that the complete MathBERT well incorporates semantic information. Besides, MathBERT retains layout structural information as well, such as the similarity scores of $ \frac{1+2}{3+4} $ and $ \frac{5+6}{7+8} $ , which are both higher than those in the former two models. The increase of $ (a+b)\times(c+d) $ in similarity score could be explained by the same substructure of $a+b$ and $c+d$. In summary, the qualitative results support that MathBERT is capable of incorporating semantic-level structural information of mathematic formulas.

\section{Conclusion}

In this paper, we propose a novel and effective pre-trained model named MathBERT, which is the first pre-trained model for mathematical formula understanding. MathBERT is jointly trained with mathematical formulas, contexts and their corresponding OPTs. The experimental results demonstrate that MathBERT achieves state-of-the-art performances on three downstream tasks including mathematical information retrieval, formula topic classification and formula headline generation. The ablation study shows that our pre-training settings could contribute to improving performance on those downstream tasks. Qualitative analysis is further conducted to show the effectiveness of MathBERT in capturing semantic-level structural information of math expressions.

\bibliographystyle{named}
\bibliography{ijcai21}

\begin{thebibliography}{}

\bibitem[\protect\citeauthoryear{Bahdanau \bgroup \em et al.\egroup
  }{2014}]{LSTM}
Dzmitry Bahdanau, Kyunghyun Cho, and Yoshua Bengio.
\newblock Neural machine translation by jointly learning to align and
  translate.
\newblock {\em arXiv preprint arXiv:1409.0473}, 2014.

\bibitem[\protect\citeauthoryear{Davila and Zanibbi}{2017}]{TangentS}
Kenny Davila and Richard Zanibbi.
\newblock Layout and semantics: Combining representations for mathematical
  formula search.
\newblock In {\em SIGIR}, pages 1165--1168, 2017.

\bibitem[\protect\citeauthoryear{Devlin \bgroup \em et al.\egroup
  }{2018}]{BERT}
Jacob Devlin, Ming-Wei Chang, Kenton Lee, and Kristina Toutanova.
\newblock Bert: Pre-training of deep bidirectional transformers for language
  understanding.
\newblock {\em arXiv preprint arXiv:1810.04805}, 2018.

\bibitem[\protect\citeauthoryear{Feng \bgroup \em et al.\egroup
  }{2020}]{codebert}
Zhangyin Feng, Daya Guo, Duyu Tang, Nan Duan, Xiaocheng Feng, Ming Gong, Linjun
  Shou, Bing Qin, Ting Liu, Daxin Jiang, and Ming Zhou.
\newblock Codebert: A pre-trained model for programming and natural languages,
  2020.

\bibitem[\protect\citeauthoryear{Jiang \bgroup \em et al.\egroup
  }{2018}]{jiang2018mathematics}
Zhuoren Jiang, Liangcai Gao, Ke~Yuan, Zheng Gao, Zhi Tang, and Xiaozhong Liu.
\newblock Mathematics content understanding for cyberlearning via formula
  evolution map.
\newblock In {\em CIKM}, pages 37--46, 2018.

\bibitem[\protect\citeauthoryear{Kristianto \bgroup \em et al.\egroup
  }{2016}]{MCAT}
Giovanni~Yoko Kristianto, Goran Topic, and Akiko Aizawa.
\newblock Mcat math retrieval system for ntcir-12 mathir task.
\newblock In {\em NTCIR}, 2016.

\bibitem[\protect\citeauthoryear{Mansouri \bgroup \em et al.\egroup
  }{2019}]{TangentCFT}
Behrooz Mansouri, Shaurya Rohatgi, Douglas~W Oard, Jian Wu, C~Lee Giles, and
  Richard Zanibbi.
\newblock Tangent-cft: An embedding model for mathematical formulas.
\newblock In {\em SIGIR}, pages 11--18, 2019.

\bibitem[\protect\citeauthoryear{Miller}{2019}]{miller2019leveraging}
Derek Miller.
\newblock Leveraging bert for extractive text summarization on lectures.
\newblock {\em arXiv preprint arXiv:1906.04165}, 2019.

\bibitem[\protect\citeauthoryear{Peters \bgroup \em et al.\egroup
  }{2018}]{ELMo}
Matthew~E Peters, Mark Neumann, Mohit Iyyer, Matt Gardner, Christopher Clark,
  Kenton Lee, and Luke Zettlemoyer.
\newblock Deep contextualized word representations.
\newblock {\em arXiv preprint arXiv:1802.05365}, 2018.

\bibitem[\protect\citeauthoryear{Radford \bgroup \em et al.\egroup
  }{2018}]{GPT}
Alec Radford, Karthik Narasimhan, Tim Salimans, and Ilya Sutskever.
\newblock Improving language understanding by generative pre-training, 2018.

\bibitem[\protect\citeauthoryear{See \bgroup \em et al.\egroup }{2017}]{PtGen}
Abigail See, Peter~J Liu, and Christopher~D Manning.
\newblock Get to the point: Summarization with pointer-generator networks.
\newblock {\em arXiv preprint arXiv:1704.04368}, 2017.

\bibitem[\protect\citeauthoryear{Sun \bgroup \em et al.\egroup
  }{2019}]{VideoBERT}
Chen Sun, Austin Myers, Carl Vondrick, Kevin Murphy, and Cordelia Schmid.
\newblock Videobert: A joint model for video and language representation
  learning.
\newblock {\em arXiv preprint arXiv:1904.01766}, 2019.

\bibitem[\protect\citeauthoryear{Sundararaman \bgroup \em et al.\egroup
  }{2019}]{sundararaman2019syntax}
Dhanasekar Sundararaman, Vivek Subramanian, Guoyin Wang, Shijing Si, Dinghan
  Shen, Dong Wang, and Lawrence Carin.
\newblock Syntax-infused transformer and bert models for machine translation
  and natural language understanding.
\newblock {\em arXiv preprint arXiv:1911.06156}, 2019.

\bibitem[\protect\citeauthoryear{Vaswani \bgroup \em et al.\egroup
  }{2017}]{Transformer}
Ashish Vaswani, Noam Shazeer, Niki Parmar, Jakob Uszkoreit, Llion Jones,
  Aidan~N Gomez, {\L}ukasz Kaiser, and Illia Polosukhin.
\newblock Attention is all you need.
\newblock In {\em Advances in neural information processing systems}, pages
  5998--6008, 2017.

\bibitem[\protect\citeauthoryear{Wang \bgroup \em et al.\egroup
  }{2015}]{wang2015wikimirs}
Yuehan Wang, Liangcai Gao, Simeng Wang, Zhi Tang, Xiaozhong Liu, and Ke~Yuan.
\newblock Wikimirs 3.0: a hybrid mir system based on the context, structure and
  importance of formulae in a document.
\newblock In {\em JCDL}, pages 173--182, 2015.

\bibitem[\protect\citeauthoryear{Xenouleas \bgroup \em et al.\egroup
  }{2019}]{xenouleas2019sumqe}
Stratos Xenouleas, Prodromos Malakasiotis, Marianna Apidianaki, and Ion
  Androutsopoulos.
\newblock Sumqe: a bert-based summary quality estimation model.
\newblock {\em arXiv preprint arXiv:1909.00578}, 2019.

\bibitem[\protect\citeauthoryear{Xu \bgroup \em et al.\egroup
  }{2020}]{LayoutLM}
Yiheng Xu, Minghao Li, Lei Cui, Shaohan Huang, Furu Wei, and Ming Zhou.
\newblock Layoutlm: Pre-training of text and layout for document image
  understanding.
\newblock {\em SIGKDD}, Jul 2020.

\bibitem[\protect\citeauthoryear{Yang \bgroup \em et al.\egroup }{2019}]{XLNet}
Zhilin Yang, Zihang Dai, Yiming Yang, Jaime Carbonell, Ruslan Salakhutdinov,
  and Quoc~V Le.
\newblock Xlnet: Generalized autoregressive pretraining for language
  understanding.
\newblock {\em arXiv preprint arXiv:1906.08237}, 2019.

\bibitem[\protect\citeauthoryear{Yasunaga and Lafferty}{2019}]{Topiceq}
Michihiro Yasunaga and John~D Lafferty.
\newblock Topiceq: A joint topic and mathematical equation model for scientific
  texts.
\newblock In {\em AAAI}, volume~33, pages 7394--7401, 2019.

\bibitem[\protect\citeauthoryear{Yuan \bgroup \em et al.\egroup
  }{2016}]{yuan2016mathematical}
Ke~Yuan, Liangcai Gao, Yuehan Wang, Xiaohan Yi, and Zhi Tang.
\newblock A mathematical information retrieval system based on rankboost.
\newblock In {\em JCDL}, pages 259--260, 2016.

\bibitem[\protect\citeauthoryear{Yuan \bgroup \em et al.\egroup
  }{2020}]{yuanke}
Ke~Yuan, Dafang He, Zhuoren Jiang, Liangcai Gao, Zhi Tang, and C~Lee Giles.
\newblock Automatic generation of headlines for online math questions.
\newblock In {\em AAAI}, pages 9490--9497, 2020.

\bibitem[\protect\citeauthoryear{Zanibbi and
  Blostein}{2012}]{zanibbi2012recognition}
Richard Zanibbi and Dorothea Blostein.
\newblock Recognition and retrieval of mathematical expressions.
\newblock {\em IJDAR}, 15(4):331--357, 2012.

\bibitem[\protect\citeauthoryear{Zanibbi \bgroup \em et al.\egroup
  }{2016}]{NTCIR-12}
Richard Zanibbi, Akiko Aizawa, Michael Kohlhase, Iadh Ounis, Goran Topic, and
  Kenny Davila.
\newblock Ntcir-12 mathir task overview.
\newblock In {\em NTCIR}, 2016.

\bibitem[\protect\citeauthoryear{Zhang \bgroup \em et al.\egroup
  }{2020}]{zhang2020tree}
Jianshu Zhang, Jun Du, Yongxin Yang, Yi-Zhe Song, Si~Wei, and Lirong Dai.
\newblock A tree-structured decoder for image-to-markup generation.
\newblock In {\em ICML}, pages 11076--11085. PMLR, 2020.

\bibitem[\protect\citeauthoryear{Zhong and Zanibbi}{2019}]{Approach0}
Wei Zhong and Richard Zanibbi.
\newblock Structural similarity search for formulas using leaf-root paths in
  operator subtrees.
\newblock In {\em European Conference on Information Retrieval}, pages
  116--129. Springer, 2019.

\bibitem[\protect\citeauthoryear{Zhu \bgroup \em et al.\egroup
  }{2020}]{BERT-fused}
Jinhua Zhu, Yingce Xia, Lijun Wu, Di~He, Tao Qin, Wengang Zhou, Houqiang Li,
  and Tie-Yan Liu.
\newblock Incorporating bert into neural machine translation.
\newblock {\em arXiv preprint arXiv:2002.06823}, 2020.

\end{thebibliography}

\end{document}